%% file: main.tex
\documentclass[11pt,a4paper]{article}
\usepackage[hyperref]{emnlp-ijcnlp-2019}
\usepackage{times}
\usepackage{latexsym}

\usepackage{microtype}

\usepackage{url}

\usepackage{graphicx}
\usepackage{arydshln}
\usepackage{caption}
\usepackage{subcaption}
\usepackage{amsmath}
\usepackage{amssymb}
\usepackage{enumitem}

\usepackage{booktabs}
\newcommand{\tabitem}{~~\llap{\textbullet}~~}

\newcommand{\Ni}{({\em i})~}
\newcommand{\Nii}{({\em ii})~}
\newcommand{\Niii}{({\em iii})~}

\aclfinalcopy 

\title{Contrastive Language Adaptation for Cross-Lingual Stance Detection}

\author{Mitra Mohtarami$^1$, James Glass$^1$, Preslav Nakov$^2$ \\
 $^1$MIT Computer Science and Artificial Intelligence Laboratory, Cambridge, MA, USA\\
  $^2$Qatar Computing Research Institute, HBKU, Doha, Qatar \\
  {\tt \{mitra,glass\}@csail.mit.edu; pnakov@hbku.edu.qa}}

\date{}

\begin{document}
\maketitle
\begin{abstract}
We study cross-lingual stance detection, which aims to leverage labeled data in one language to identify the relative perspective (or stance) of a given document with respect to a claim in a different target language. In particular, we introduce a novel contrastive language adaptation approach applied to memory networks, which ensures accurate alignment of stances in the source and target languages, and can effectively deal with the challenge of limited labeled data in the target language. The evaluation results on public benchmark datasets and comparison against current state-of-the-art approaches demonstrate the effectiveness of our approach.
\end{abstract}

\input{introduction}
\input{CLMN}
\input{results}
\input{discussions}

\input{related_work}

\input{conclusion}

\section*{Acknowledgments}

We thank the anonymous reviewers for their insightful comments. This research was supported in part by the Qatar Computing Research Institute, HBKU\footnote{This research is part of the Tanbih project (available at \url{http://tanbih.qcri.org/}) which aims to limit the effect of ``fake news'', propaganda and media bias by making users aware of what they are reading.} and DSTA of Singapore.

\bibliography{emnlp-ijcnlp-2019}
\bibliographystyle{acl_natbib}

\end{document}

%% file: introduction.tex
\section{Introduction}
\label{sec:introduction}


The rise of social media has enabled the phenomenon of ``fake news,'' which could target specific individuals and can be used for deceptive purposes~\cite{Lazer1094,Vosoughi1146}. 
As manual fact-checking is a time-consuming and tedious process, computational approaches have been proposed as a possible alternative~\cite{Popat:2017:TLE:3041021.3055133,P17-2067,AAAIFactChecking2018}, based on information sources such as social media~\cite{P17-1066}, Wikipedia~\cite{thorne-EtAl:2018:N18-1}, and knowledge bases~\cite{Huynh:2018:TBF:3184558.3191616}.
%
Fact-checking 
is a multi-step process
\cite{vlachos2014fact}:
(\emph{i})~checking the reliability of media sources, 
(\emph{ii})~retrieving potentially relevant documents from reliable sources as evidence for each target claim,
(\emph{iii})~predicting the stance of each document with respect to the target claim, and finally
(\emph{iv})~making a decision based on the stances from (\emph{iii}) for all documents from (\emph{ii}). 
%

Here, we focus on stance detection which aims to identify the relative perspective of a document with respect to a claim, typically modeled using labels such as \emph{agree}, \emph{disagree}, \emph{discuss}, and \emph{unrelated}.

\noindent Current approaches to stance detection~\cite{Bar-Haim,Dungs-Sebastian,Kochkina-Elena,Inkpen, mitra2018memory} are well-studied in mono-lingual settings, in particular for English, but less attention has been paid to other languages 
and {\em cross-lingual} settings. This is partially due to domain differences and to the lack of training data in other languages. 

We aim to bridge this gap by proposing a cross-lingual model for stance detection. Our model leverages resources of a source language (e.g., English) to train a model for a target language (e.g., Arabic). 
Furthermore, we propose a novel contrastive language adaptation approach that effectively aligns samples with similar or dissimilar stances across source and target languages using task-specific loss functions. We apply our language adaptation approach to memory networks~\cite{NIPS2015_5846}, which have been found effective for mono-lingual stance detection~\cite{mitra2018memory}.

Our model can explain its predictions about stances of documents against claims in a different/target language by extracting relevant text snippets from the documents of the target language as evidence. We use evidence extraction as a measure to evaluate the
trasferability of our model. This is because more accurate evidence extraction indicates that the model can better learn semantic relations between claims and pieces of evidence, and consequently can better transfer knowledge to the target language.

The contributions of this paper are summarized as follows:
\begin{itemize}[itemsep=1pt,topsep=1pt]
\item We propose a novel language adaptation approach based on contrastive stance alignment that aligns the class labels between source and target languages for effective cross-lingual stance detection.
\item Our model is able to extract accurate text snippets as evidence to explain its predictions in the target language (results are in Section~\ref{dis:Transferrability}). 
\item To the best of our knowledge, this is the first work on cross-lingual stance detection.
%
\end{itemize}

We conducted our experiments on English (as source language) and Arabic (as target language). In particular, we used the Fake News Challenge dataset~\cite{Retrospective-C18-1158} as source data and an Arabic benchmark dataset~\cite{baly2018integrating} as target data. The evaluation results have shown $2.7$ and $4.0$ absolute improvement in terms of macro-F1 and weighted accuracy for stance detection over the current state-of-the-art mono-lingual baseline, and $11.4$, $14.9$, $16.1$, $12.9$, and $13.1$ points of absolute improvement in terms of precision at ranks $1$--$5$ for extracting evidence snippets respectively. Furthermore, a key finding in our investigation is that, in contrast to other tasks~\cite{devlin2018bert,peters-EtAl:2018:N18-1}, pre-training with large amounts of source data is less effective for cross-lingual stance detection. We show that this is because pre-training can considerably bias the model toward the source language.

%% file: CLMN.tex
\begin{figure*}
\centering
\includegraphics[scale=0.12]{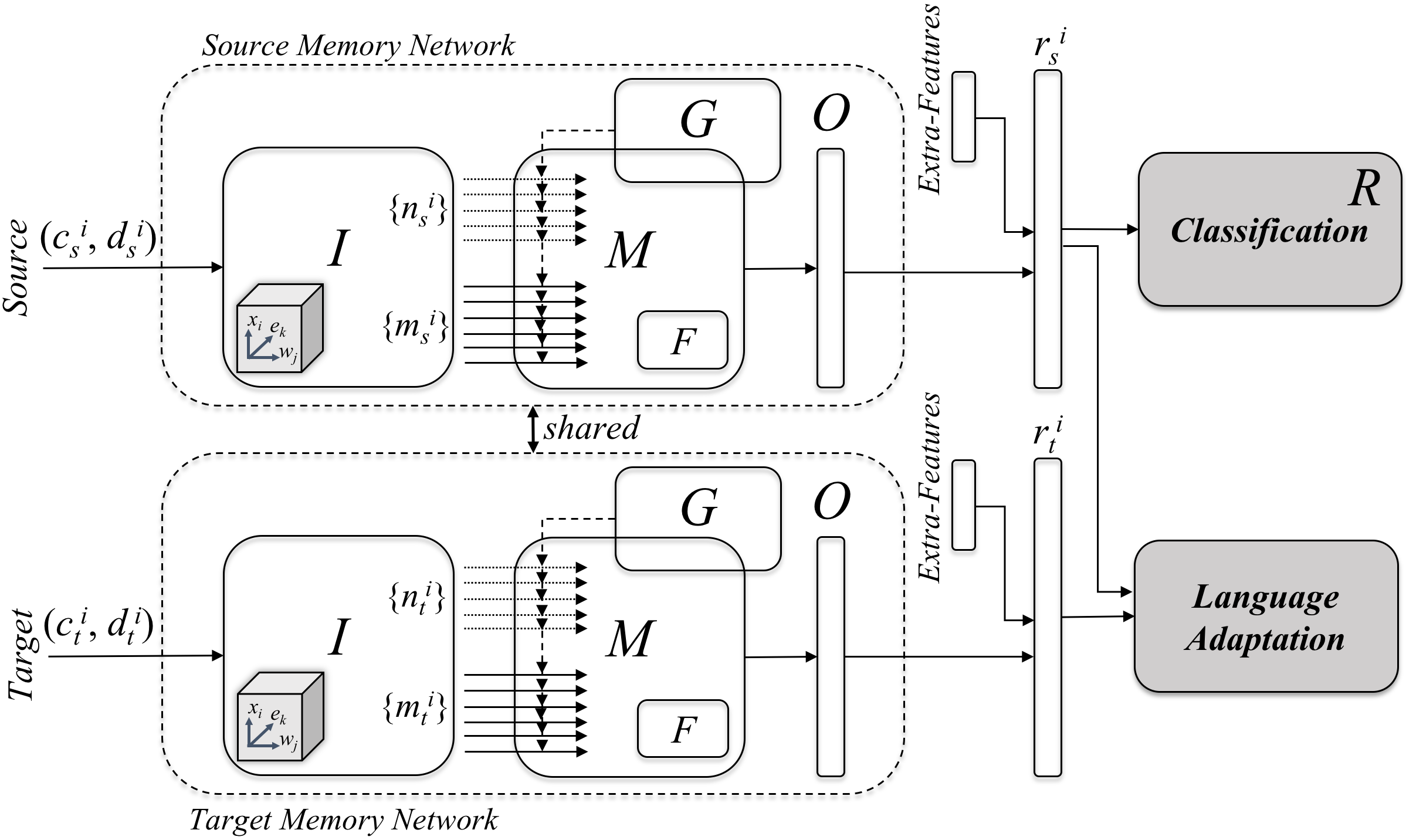}
\caption{The architecture of our cross-lingual memory network for stance detection.}
\label{fig:LAMN}
\end{figure*}

\section{Method}
\label{sec:CLMN_stance}

Assume that we are given a training dataset for a source language, $\mathcal{D}_s$, which contains a set of triplets as follows: $\mathcal{D}_s=\big\{\big((c_i^s,d_i^s), y_i^s\big)\big\}_{i=1}^N$, where $N$ is the number of source samples, $(c_i^s,d_i^s)$ is a pair of claim $c_i^s$ and document $d_i^s$, and $y_i^s\in Y$, $Y = \{\textit{agree}, \textit{disagree}, \textit{discuss}, \textit{unrelated}\}$, is the corresponding label indicating the stance of the document with respect to the claim. In addition, we are given a very small training dataset for the target language $\mathcal{D}_t=\big\{\big((c_i^t,d_i^t), y_i^t\big)\big\}_{i=1}^M$, where $M$ ($M<<N$) is the number of target samples, and $(c_i^t,d_i^t)$ is a pair of claim and document in the target language with stance label $y_i^t\in Y$. In reality, \Ni~the size of the target dataset is very small, 
\Nii~claims and documents in the source and target languages are from different domains, and \Niii~the only commonality between the source and target datasets is in their stance labels, i.e., $y_i^s, y_i^t \in {Y}$. 

We develop a language adaptation approach to effectively use the commonality between the source and the target datasets in their label space and to deal with the limited size of the target training data. We apply our language adaptation approach to end-to-end memory networks~\cite{NIPS2015_5846} for cross-lingual stance detection. 

\noindent We use memory networks as they have achieved state-of-the-art performance for mono-lingual stance detection~\cite{mitra2018memory}. However, our language adaptation approach can be applied to any other type of neural network. 
The architecture of our cross-lingual stance detection model is shown in Figure~\ref{fig:LAMN}. It has two main components: (\emph{i})~\emph{Memory Networks} indicated with two dashed boxes for the source and the target languages, and (\emph{ii})~\emph{Contrastive Language Adaptation} component. In what follows, we first explain our memory network model for cross-lingual stance detection (Section~\ref{sec:MN}) and then present our contrastive language adaptation approach (Section~\ref{sec:CLA}).  

\subsection{Memory Networks}\label{sec:MN}
Memory networks are designed to remember past information~\cite{NIPS2015_5846} and have been successfully applied to NLP tasks ranging from dialog~\cite{BordesW16} to question answering~\cite{Xiong2016DMN} and mono-lingual stance detection~\cite{mitra2018memory}. 
They include components that can potentially use different learning models and inference strategies. 
Our source and target memory networks follow the same architecture as depicted in Figure~\ref{fig:LAMN}:

A memory network consists of six components. The network takes as input a document $d$ and a claim $c$ and encodes them into the {\em input space} $I$. These representations are stored in the {\em memory} component $M$ for future processing. The relevant parts of the input are identified in the {\em inference} component $F$, and used by the {\em generalization} component $G$ to update the memory $M$. Finally, the {\em output} component $O$ generates an output from the updated memory, and encodes it to a desired format in the {\em response} component $R$ using a prediction function, e.g., \texttt{softmax} for classification tasks. We elaborate on these components below.

\paragraph{Input representation component $I$:}
It encodes documents and claims into corresponding representations.
%
Each document $d$ is divided into a sequence of paragraphs $X = (x_1,\dots, x_l)$, where each $x_j$ is encoded as $\mathbf{m}_j$ using an LSTM network, and as $\mathbf{n}_j$ using a CNN; these representations are stored in the memory component $M$. Note that while LSTMs
are designed to capture and memorize their inputs~\cite{acl:TanSXZ16}, CNNs emphasize the local interaction between individual words in sequences, which is important for obtaining good representation~\cite{kim:2014:EMNLP2014}. 

\noindent Thus, our $I$ component uses both LSTM and CNN representations. It also uses separate LSTM and CNN with their own parameters to represent each input claim $c$ as $\mathbf{c}_{\textit{lstm}}$ and $\mathbf{c}_\textit{cnn}$, respectively.

We consider each paragraph as a single piece of {\em evidence} because a paragraph usually represents a coherent argument, unified under one or more inter-related topics. We thus use the terms paragraph and evidence interchangeably.

\paragraph{Inference component $F$:}
Our inference component computes LSTM- and CNN-based similarity between each claim $c$ and evidence $x_j$ as follows:
\begin{align*}\small
P^{\textit{lstm}}_j &= {\mathbf{c}_{\textit{lstm}}}^\intercal\times\mathbf{M}\times \mathbf{m}_j, \forall{j} \\
P^{\textit{cnn}}_j &=  {\mathbf{c}_{\textit{cnn}}}^\intercal\times\mathbf{M'}\times \mathbf{n}_j, \forall{j}
\end{align*}
where 
$P^{\textit{lstm}}_.$ and $P^{\textit{cnn}}_.$ indicate claim-evidence similarity based on LSTM and CNN respectively, 
$\mathbf{c}_{\textit{lstm}}\in\mathbb{R}^q$ and $\mathbf{m}_j\in\mathbb{R}^d$ are LSTM representations of $c$ and $x_j$ respectively, $\mathbf{c}_{\textit{cnn}}\in\mathbb{R}^{q'}$ and $\mathbf{n}_j\in\mathbb{R}^{d'}$ are the corresponding CNN representations, and
$\mathbf{M}\in\mathbb{R}^{q\times d}$ and $\mathbf{M}'\in\mathbb{R}^{q'\times d'}$ are similarity matrices trained to map 
claims and paragraphs into the same space with respect to their LSTM and CNN representations.
The rationale behind using these similarity matrices is that, in the memory network, we seek a transformation of the 
input claim, i.e.,~$\mathbf{c}^\intercal\times\mathbf{M}$, in order to obtain the closest evidence to the claim.

Additionally, we compute another semantic similarity vector, $P^{\textit{tfidf}}_j$ , by applying a cosine similarity between the TF.IDF~\cite{sparck2004idf} representation of $x_j$ and $c$. This is particularly useful for stance detection as it can help filtering out unrelated pieces of evidence.

\paragraph{Memory $M$ and Generalization $G$ components:}
Our memory component stores representations and the generalization component improves their quality by filtering out unrelated evidence. For example, the LSTM representations of paragraphs, $\mathbf{m}_j, \forall{j}$, are updated using the claim-evidence similarity $P^{\textit{tfidf}}_j$ as follows: $ \mathbf{m}_j = \mathbf{m}_j\odot P^{\textit{tfidf}}_j, \forall{j}$. 
This transformation will help filter out unrelated evidence with respect to claims. 
The updated $\mathbf{m}_j$ in conjunction with $\mathbf{c}_{\textit{lstm}}$ are used by the inference component $F$ to 
compute $P^{\textit{lstm}}_j, \forall{j}$ as we explained above.
Then, $P^{\textit{lstm}}_j$ are in turn used to update CNN representations in memory as follows: $\mathbf{n}_j =  \mathbf{n}_j\odot P^{\textit{lstm}}_j, \forall{j}$. 
Finally, the updated $\mathbf{n}_j$ and $\mathbf{c}_{\textit{cnn}}$ are used to compute $P^{\textit{cnn}}_j$.

\paragraph{Output representation component $O$:}
This component computes the output of the memory $M$ by concatenating
the average vector of the updated $\mathbf{n}_j$ with the {\it maximum} and {\it average} of claim-evidence similarity vectors $P^{\textit{tfidf}}_j$, $P^{\textit{lstm}}_j$ and $P^{\textit{cnn}}_j$. 
The maximum helps to identify parts of documents that are most similar to claims, while the average estimates the overall document-claim similarity.

\paragraph{Response generation component $R$:}
This component computes the final stance of a document with respect to a claim.
For this, the output of component $O$ is concatenated with $\mathbf{c}_\textit{lstm}$ and $\mathbf{c}_\textit{cnn}$ and fed into a \texttt{softmax} to predict the stance of the document with respect to the claim.
	
All the memory network parameters, including those of CNN and LSTM in the $I$ component, the similarity matrices $\mathbf{M}$ and $\mathbf{M}'$ in $F$, and
the classifier parameters in $R$, are jointly learned during the training process with our language adaptation.


\subsection{Contrastive Language Adaptation}\label{sec:CLA}

\begin{figure}[t]
\centering
\includegraphics[width=0.65\linewidth]{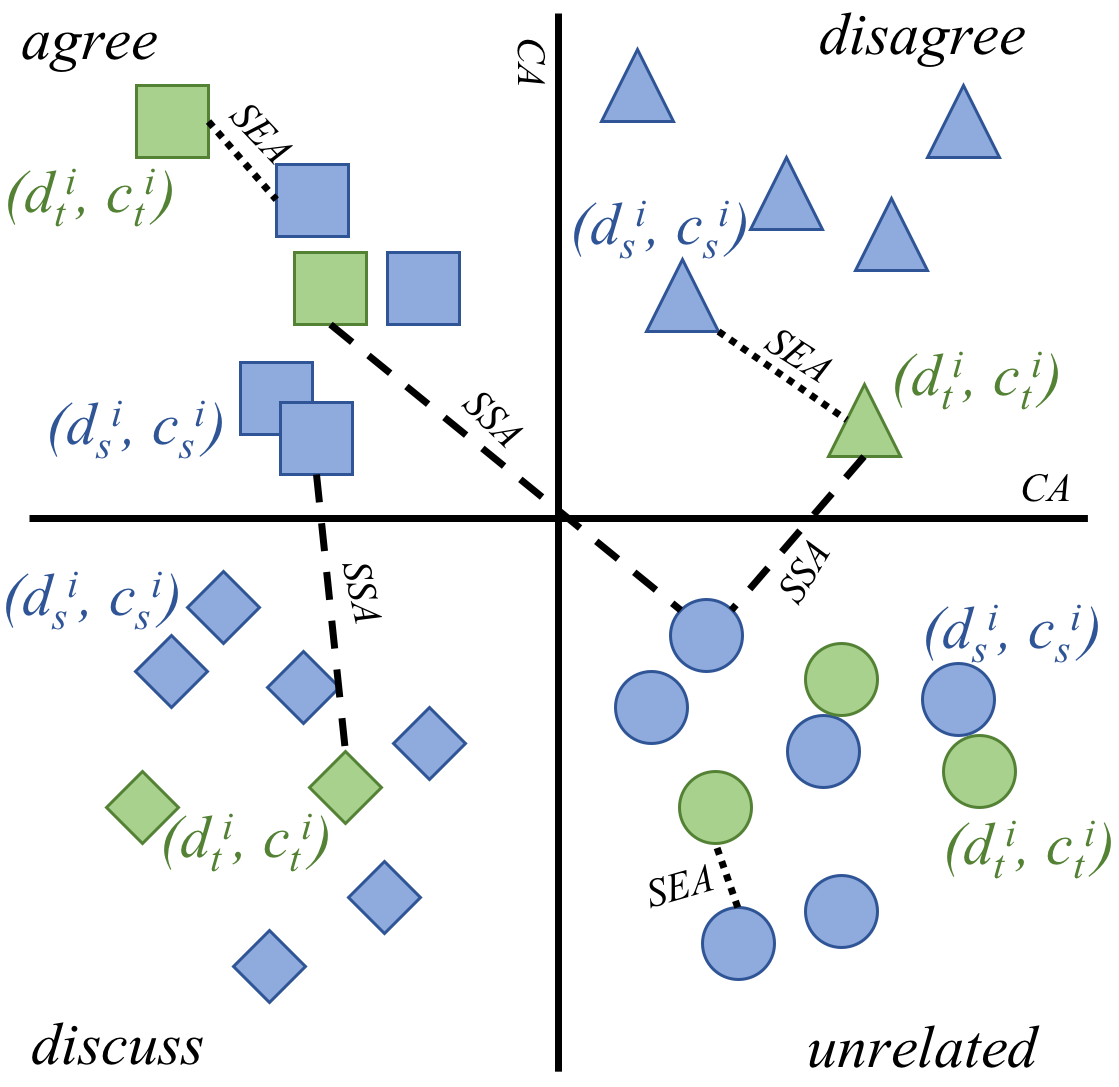}
\caption{Illustration of stance equal alignment (SEA), stance separation alignment (SSA), and classification alignment (CA) constraints. Different shapes indicate different stance labels and colors specify source (\textit{blue}) and target (\textit{green}) languages.}
\label{fig:alignments}
\vspace{-0.05in}
\end{figure}

\begin{table}[t]
\small
\centering
\begin{tabular}{p{.1cm}p{6.5cm}}\\\hline
\multicolumn{2}{@{}p{7cm}}{{\bf Algorithm 1. Cross-Lingual Stance Detection Model}} \\\hline
\multicolumn{2}{@{}p{7cm}}{{\bf Inputs:}}\\
\tabitem  & $\{(d_i^s, c_i^s), y_i^s\}_1^{l^s}$: set of pairs of documents $d_i^s$ and claims $c_i^s$ in the source language, where $y_i^s \in Y$, $Y = \{agree, disagree, discuss, unrelated\}$\\
\tabitem & $\{(d_i^t, c_i^t),y_i^t\}_1^{l^t}$: small set of pairs of documents $d_i^t$ and claims $c_i^t$ in the target language with $y_i^t \in Y$\\
\multicolumn{2}{@{}p{7cm}}{{\bf Output:}} \\
\tabitem & $\{(d_i^t, c_i^t) \rightarrow y_i^t\}_{l^t+1}^M$: assign stance labels $y_i^t \in Y$ to given unlabeled target pairs\\
\multicolumn{2}{@{}p{7cm}}{{\bf Cross-Lingual model:}} \\
$1$ & Create the sets $\{\big((d_i^s, c_i^s), (d_j^t, c_j^t)\big),(y_i^s,y^{\prime}_i)\}$ and 
$\{\big((d_j^t, c_j^t), (d_i^s, c_i^s)\big),(y_j^t,y^{\prime}_j)\}$, where $y^{\prime} =1$ if the source and the target have the same label, and $y^{\prime} =0$ otherwise.\\
$2$ & \bf{Loop for $e$ epochs:} \\
$3$ & \hspace{5pt} pass $(d_i^s, c_i^s)$ to the source memory network to\\
    & \hspace{5pt} create its representation $r_i^s$.\\
$4$ & \hspace{5pt} pass $(d_j^t, c_j^t)$ to the target memory network to\\
    & \hspace{5pt} create its representation $r_j^t$.\\
$5$ & \hspace{5pt} pass $(r_i^s, y_i^s)$ to the classification to compute its\\
    & \hspace{5pt} classification loss ${L_{CA}}_i^s$.\\
$6$ & \hspace{5pt} pass $(r_i^s, r_i^t, y^{\prime i})$ to the language adaptation to\\
    & \hspace{5pt} compute the stance alignment loss ${L_{CSA}}_i$. \\
$7$ & \hspace{5pt} compute total loss $L_i^s = (1-\alpha) {L_{CA}}_i^s + \alpha {L_{CSA}}_i$.\\
$8$ & repeat steps $2$-$7$ with a change in step $5$ by passing the target sample $(r_j^t, y_j^t)$ to the classification instead of the source sample, and compute its $L_j^t$ in step $7$.\\
$9$ & jointly optimize all parameters of the model using the average loss $L = \mathrm{mean}(\{L_i^s\}+\{L_j^t\})$.\\
\hline
\end{tabular}
\caption{Cross-lingual stance detection model.}
    \label{tbl:alg_LAMN}
\end{table}

Memory networks are effective for stance detection in mono-lingual settings~\cite{mitra2018memory} when there is sufficient training data. However, we show that these models have limited transferability to target languages with limited data. This could be due to discrepancy between the underlying data distributions in the source and target languages.
We show that the performance of these networks can be trivially increased when the model, pre-trained on source data, is fine-tuned using small amounts of labeled target data. 
We further develop contrastive language adaptation that can exploit the labeled source data to perform well on target data. 

\noindent Our contrastive adaptation approach:
\begin{itemize}[noitemsep,topsep=0pt]
\item encourages pairs $( d_i^s, c_i^s)$ from the source language and $( d_i^t, c_i^t)$ from the target language with the same stance labels (i.e., $y_i^s = y_i^t$) to be nearby in the embedding space. We call this mapping \textit{Stance Equal Alignment} (SEA), illustrated with dotted lines in Figure~\ref{fig:alignments}. Note that documents and claims in the two languages are often semantically different and are not corresponding translations of each other.

\item encourages pairs $( d_i^s, c_i^s)$ from the source language and $( d_i^t, c_i^t)$ from the target language with different stance labels (i.e., $y_i^s \neq y_i^t$) to be far apart in the embedding space. We call this mapping \textit{Stance Separation Alignment} (SSA), shown with dashed lines in Figure~\ref{fig:alignments}.

\item encourages pairs $( d_i^s, c_i^s)$ from the source language and $( d_i^t, c_i^t)$ from the target language to be correctly classified as $( d_i^s, c_i^s) \rightarrow y_i^s$ and $( d_i^t, c_i^t) \rightarrow y_i^t $. We call this \textit{Classification Alignment} (CA), solid lines in Figure~\ref{fig:alignments}.
\end{itemize}

\noindent We make complete use of the stance labels in the cross-lingual setting by parameterizing our model according to the distance between the source and the target samples in the embedding space.

For stance equal alignment (SEA) constraint, the objective is to 
minimize the distance between pairs of source and target data with the same stance labels. We achieve this using the following loss:
\begin{equation}\label{eq:L_SEA}\small
  L_{SEA}=\begin{cases}
    \texttt{D}\big(\texttt{g}(d_i^s,c_i^s) - \texttt{g}(d_i^t,c_i^t)\big)^2, & y_i^s=y_i^t\\
    0, & \text{otherwise},
  \end{cases}
\end{equation}
where \texttt{g} maps its input pair to an embedding space using our memory network or any mono-lingual model, and \texttt{D} computes Euclidean distance.

For stance separation alignment (SSA), the goal is to maximize the distance between pairs with different stance labels. We use the following loss:
\begin{equation}\label{eq:L_SSA}\small
  L_{SSA}=\begin{cases}
    \max\Big( 0, m-\\
       \hspace{20pt}\texttt{D}\big( \texttt{g}(d_i^s,c_i^s) - \texttt{g}(d_i^t,c_i^t) \big)^2 \Big), & y_i^s\neq y_i^t\\
    0, & \text{otherwise},
  \end{cases}
\end{equation}
where  we maximize the distance between pairs with different stance labels up to the margin $m$.

\noindent The margin parameter $m$ specifies the extent of separability in the embedding space. 

We can further use any classification loss to enforce classification alignment (CA). We use categorical cross-entropy and call it \emph{Classification Alignment} loss $L_{CA}$.

We develop our overall language adaptation loss, named \textit{Contrastive Stance Alignment} loss, $L_{CSA}$, by combining $L_{SEA}$ and $L_{SSA}$ as follows:
\begin{equation}\label{eq:L_CSA}\small
L_{CSA} = L_{SEA} + L_{SSA}.
\end{equation}
Finally, the total loss of our cross-lingual stance detection model is defined as follows:
\begin{equation}\label{eq:L_total}\small
L = (1-\alpha) L_{CA} + \alpha L_{CSA},
\end{equation}
where the $\alpha$ parameter controls the balance between classification and language adaptation losses, which we optimize on the validation dataset. 

\paragraph{Information Flow:}
Our overall cross-lingual model for stance detection is shown in Figure~\ref{fig:LAMN}, and a summary of the algorithm is presented in Table~\ref{tbl:alg_LAMN}. As Figure~\ref{fig:LAMN} shows, each source and target pairs are passed to the source and to the target memory networks to obtain their corresponding representations (Lines $3$-$4$ in Table~\ref{tbl:alg_LAMN}). The source representation and its gold stance label are 
passed to the classifier to compute the classification loss (Line $5$). In addition, the source and the target representations in conjunction with a binary parameter ($y^{\prime}$, which is $1$ if the source and the target have the same stance label, and $0$ otherwise) are passed to the language adaptation component to compute the contrastive stance alignment loss $L_{CSA}$ (Line $6$). Finally, the total loss is computed based on Equation~(\ref{eq:L_total}) (Line $7$).

The classifier also uses labeled target samples to create a shared embedding space and to fine-tune itself with respect to the target language. 
For this purpose, we repeat the above steps by switching the target and the source pipelines (Line $8$). Finally, we compute the average of all losses and we use it to optimize the parameters of our model (Line $9$).



\paragraph{Pre-training for Language Adaptation:}
Pre-training has been found effective in many language adaptation settings~\cite{Adda_CVPR2017}.
To investigate the effect of pre-training, we first pre-train the source memory network and the classifier using $\mathcal{D}_s$ (only the top pipeline in Figure~\ref{fig:LAMN}), and then we apply language adaptation with the full model. 

%% file: results.tex

\begin{table*}[t]
\centering
\scalebox{0.85}{
\begin{tabular}{p{5.58cm}@{ }|ccc:cc}
\hline
  \bf Methods & \bf Weigh. Acc. & \bf Acc. & \bf Macro-F1 & \bf F1 ({\it agree, disagree, discuss, unrelated}) \\ \hline
1. \ \ \ All-\emph{unrelated} & 34.8 & 68.1 & 20.3 & 0 / 0 / 0 / 81.0 \\
2. \ \ \ All-\emph{agree} & 40.2 & 15.6 & 6.7 & 27.0 / 0 / 	0 / 0 \\
\hdashline[1.5pt/2pt]
3. \ \ \ Gradient Boosting{\tiny~\cite{baly2018integrating}} & 55.6 & \bf 72.4 & 41.0 & 60.4 / 9.0 / 10.4 / 84.0 \\
4. \ \ \ TFMLP{\tiny~\cite{riedel2017simple}} & 49.3 & 66.0 & 37.1 & 47.0 / 7.8 / 13.4 / 80.0 \\
5. \ \ \ EnrichedMLP{\tiny~\cite{Retrospective-C18-1158}} & 55.1 & 70.5 & 41.3 & 59.1 / 9.2 / 14.1 / 82.3 \\
6. \ \ \ Ensemble {\tiny~\cite{baird2017talos}} & 53.6 & 71.6 & 37.2 & 57.5 / 2.1 / 6.2 / 83.2 \\
\hline
7. \ \ \ MN {\it \tiny $target \rightarrow target$}{\tiny~\cite{mitra2018memory}} & 55.3 & 70.9 & 41.7 & 60.0 / 15.0 / 08.5 / 83.1\\
8. \ \ \ MN {\it \tiny $source \rightarrow target$} & 53.2 & 64.2 & 36.0 & 40.4 / 02.0 / 19.1 / 82.4\\
9. \ \ \ MN {\it \tiny $(source, target) \rightarrow target$} & 57.3 & 65.0 & 42.5 & 58.0 / 12.2 / 20.9 / 79.0\\
\hline
10.\ \ ADMN {\it \tiny (adversarial)} & 58.6 & 58.4 & 43.4 & 60.2 / 16.1 / 24.2 / 72.9\\  
\hline
11.\ \ CLMN {\it \tiny (contrastive)} & \bf 61.3 & \underline{71.6} & \bf 45.2 & 65.1 / 11.6 / 20.5 / 83.7\\
\hline
\end{tabular}}
\caption{\label{tbl:results}Evaluation results on the target Arabic test dataset.}
\end{table*}

\section{Experiments}\label{sec:experiments}

\paragraph{Data and Settings.}
As source data, we use the Fake News Challenge dataset\footnote{Available at \url{www.fakenewschallenge.org}} which contains $75.4$K claim-document pairs in English with \{\textit{agree}, \textit{disagree}, \textit{discuss}, \textit{unrelated}\} as stance labels. As target dataset, we use $3$K Arabic claim-document pairs developed in~\cite{baly2018integrating}.\footnote{Available at \url{http://groups.csail.mit.edu/sls/downloads/}}

We perform $5$-fold cross-validation on the Arabic dataset, using each fold in turn for testing, and keeping $80$\% of the remaining data for training and $20$\% for development.
We use $300$-dimensional pretrained cross-lingual Wikipedia word embeddings from MUSE~\cite{conneau2017word}. 
We use $300$-dimensional units for the LSTM and $100$ feature maps with filter width of $5$ for the CNN. We consider the first $9$ paragraphs per document, which is the median number of paragraphs in source documents.
We optimize all hyper-parameters on validation data using \emph{Adam}~\cite{corr:KingmaB14}. 

\paragraph{Evaluation Measures.}
We use the followings:

\begin{itemize}[noitemsep,topsep=0pt]

\item \emph{Accuracy}: The fraction of correctly classified examples. 
For multi-class classification, accuracy is equivalent to micro-average F$_1$~\cite{Manning:2008:IIR:1394399}. 

\item \emph{Macro-F$_1$}: The average of F$_1$ scores that were calculated for each class separately.

\item {\it Weighted Accuracy}: 
This is a hierarchical metric, which first awards $0.25$ points if the model correctly predicts a document-claim pair as \emph{related}\footnote{\emph{related} = \{\emph{agree}, \emph{disagree}, \emph{discuss}\}} or \emph{unrelated}. If it is \emph{related}, $0.75$ additional points are assigned if the model correctly predicts the pair as \emph{agree}, \emph{disagree}, or \emph{discuss}. The goal of this weighting schema is to balance out the large number of \emph{unrelated} examples~\cite{Retrospective-C18-1158}.

\end{itemize}


\paragraph{Baselines.}

We consider the following baselines:
\begin{itemize}[noitemsep,topsep=0pt]

\item {\bf Heuristic}: Given the imbalanced nature of our data, we use two heuristic baselines where all test examples are labeled as \emph{unrelated} or \emph{agree}. The former is a majority class baseline favoring accuracy and macro-F$_1$, while the latter is better for weighted accuracy.  

\item {\bf Gradient Boosting}~\cite{baly2018integrating}: This is a Gradient Boosting classifier with $n$-gram features as well as indicators for refutation and polarity.

\item {\bf TFMLP}~\cite{riedel2017simple}: This is an MLP with normalized bag-of-words features enriched with a single TF.IDF-based similarity feature for each claim-document pair.

\item {\bf EnrichedMLP}~\cite{Retrospective-C18-1158}: This model combines five MLPs, each with six hidden layers and advanced features from topic models, latent semantic analysis, etc. 

\item {\bf Ensemble}~\cite{baird2017talos}: 
It is an ensemble based on weighted average of a deep convolutional neural network and a gradient-boosted decision tree - the best model at the Fake News Challenge. 

\item {\bf Mono-lingual Memory Network}~\cite{mitra2018memory}: This model is the current state-of-the-art for stance detection on our source dataset. It is an end-to-end memory network which incorporates both CNN and LSTM for prediction.

\item {\bf Adversarial Memory Network}:
We use adversarial domain adaptation~\cite{Ganin:2016} instead of contrastive language adaptation in our cross-lingual memory network.

\end{itemize}

\paragraph{Results.}

Table~\ref{tbl:results} shows the performance of all models on the target \textit{Arabic} test set. 
The All-\emph{unrelated} and All-\emph{agree} baselines perform poorly across evaluation measures; All-\emph{unrelated} performs better than All-\emph{agree} because \emph{unrelated} is the dominant class ($\sim68$\% of examples).

Rows $3$--$6$ show that \emph{Gradient Boosting} and \emph{EnrichedMLP} yield similar results, while \emph{TFMLP} performs the worst. We attribute this to the advanced features used in the two former models. \emph{Gradient Boosting} has better accuracy due to its better performance on the dominant class. Note that \emph{Ensemble} performs poorly because of the limited labeled data, which is insufficient to train a good CNN model.

\noindent Rows $7$--$9$ show the results for the mono-lingual memory network (MN) from~\cite{mitra2018memory}. The performance of this model when trained on Arabic data only (row $7$) is comparable to previous baselines (rows $3$--$6$). But, it shows poor performance if trained on source English data and tested on Arabic test data (row $8$). 
The model performs best (in terms of weighted accuracy and F1) if first pretrained on source data and then fine-tuned on target training data (row 9). 

Row $10$ in Table~\ref{tbl:results} shows the results for adversarial memory network (ADMN). It improves the performance of mono-lingual MN on weighted accuracy and F1, but its accuracy significantly drops. This is because adversarial approaches give higher weights to samples of the majority class (i.e., \textit{unrelated}) which makes classification more challenging for the discriminator~\cite{DBLP:journals/corr/abs-1811-08812}.

Row $11$ shows the results for our cross-lingual memory network (CLMN)
with ($\alpha=.7$); $\alpha$
controls the balance between classification and language adaptation losses (tuned using validation data).
CLMN outperforms other baselines in terms of weighted accuracy and F1 while showing comparable accuracy. We show that the improvement is due to language adaptation being able to effectively transfer knowledge from the source to target language (see Section~\ref{dis:Transferrability}).

The last column in Table~\ref{tbl:results} shows that \emph{unrelated} examples are the easiest ones.
Also, although the \emph{agree} and the \emph{discuss} classes have roughly the same size, i.e., $474$ and $409$ examples, respectively, the results for \emph{agree} are notably higher. This is mainly because the documents that discuss a claim often share the same topic with the claim, but they do not take a stance.
In addition, the \emph{disagree} examples are the most difficult ones;
this class is by far the smallest one, with only $87$ examples.

%% file: discussions.tex
\begin{figure}[t]
\centering
\includegraphics[width=0.9\linewidth]{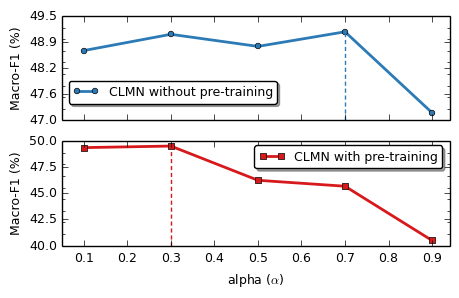}
\vspace{-7pt}
\caption{Impact of pretraining on the macro-F1 across $\alpha$ values. The Y axes show average 
results on the validation datasets with $5$-fold cross-validation.}
\label{fig:LAMN_with_without_pretraining}
\end{figure}

\begin{table}[t]\small
\centering
\scalebox{0.85}{
\begin{tabular}{l|ccc}
\hline
 \bf Methods & \bf Weigh. Acc. & \bf Acc. & \bf Macro-F1\\ \hline
CLMN {\it \tiny (with pretraining)} & 60.2 & 69.8 & 43.2\\
CLMN {\it \tiny (without pretraining)} & \bf 61.3 & \bf{71.6} & \bf 45.2 \\
\hline
\end{tabular}}
\caption{CLMN results on the target test dataset.}
\label{tbl:CLMN-Results}
\end{table}

\section{Discussion}\label{sec:discussions}

\subsection{Effect of Pretraining}\label{dis:pretraining}
Table~\ref{tbl:CLMN-Results} shows CLMN without pretraining ($\alpha=.7$) performs better on target test data than CLMN with pretraining ($\alpha=.3$), recall that $\alpha$ controls the balance between classification and language adaptation losses. Our further analysis shows that pretraining biases the model toward the source language.
Figure~\ref{fig:LAMN_with_without_pretraining} shows the impact of 
pretraining on macro-average F1 score for CLMN across different values of $\alpha$ on validation data. While the model without pretraining  achieves its best performance with a large $\alpha$ ($\alpha=0.7$), the model with pretraining performs well with a smaller $\alpha$ ($\alpha=0.3$). 
This suggests that our model can capture the characteristics of the source dataset via pretraining when using small supervision from language adaptation (i.e., small $\alpha$). However, pretraining introduces bias to the source space and the performance drops when larger weights are given to language adaptation; see the results with pretraining in Figure~\ref{fig:LAMN_with_without_pretraining}.

\subsection{Assessment of Model Transferability}\label{dis:Transferrability}

The improvements of CLMN model over the monolingual MN models that use the target only, the source only,
or both the target and the source (rows $7$--$9$ in Table~\ref{tbl:results} respectively) 
indicate its transferability.
We further estimate transferability by measuring the accuracy of the models in extracting 
evidence that support their predictions. A more accurate model should better 
transfer knowledge to the target language by accurately 
learning the relations between claims and pieces of evidence. 

\noindent Our target data has annotations (in terms of binary labels) for each piece of evidence (here paragraph) that indicate whether 
it is a rationale for the \textit{agree} or for the \textit{disagree} class.
Moreover, our inference component ($I$) has a claim-evidence similarity vector, 
$P^{cnn}_{j}$, which can be used to rank pieces of evidence from the target document against the target claim. 

\begin{figure}[t]
\centering
\includegraphics[width=0.9\linewidth]{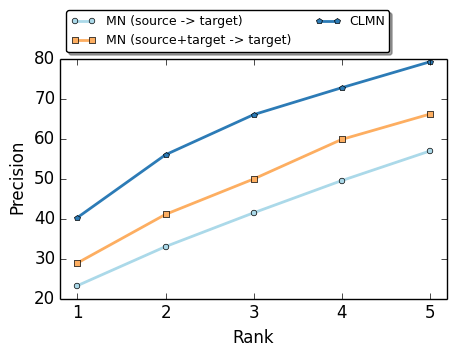}
\caption{Transferability of our cross-lingual model.}
\label{fig:transferability}
\vspace{-5pt}
\end{figure}

We use the gold data and the rankings produced by our model in order to measure its precision in extracting evidence that supports its predictions.
Figure~\ref{fig:transferability} shows that our CLMN model achieves precision of $40.2$, $55.9$, $66.0$, $72.7$, and $79.2$ at ranks $1$--$5$ respectively, and outperforms mono-lingual MN models. This indicates that CLMN can better generalize and transfer knowledge 
to the target language through learning relations between pieces of evidence and claims.

\subsection{Effect of Language Adaptation}\label{dis:Loss-Trends}

\begin{figure}[t]
\centering
\begin{subfigure}[b]{0.22\textwidth}
\centering
\includegraphics[scale=.33]{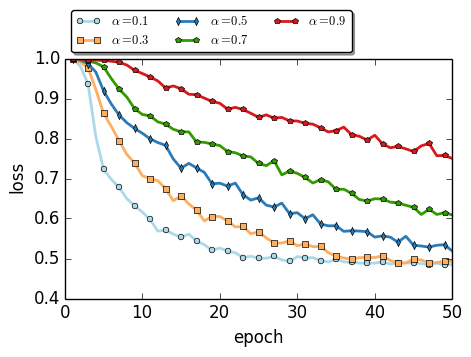} 
\caption{Classification loss}
\label{fig:Classification_loss_throughAlphas}
\end{subfigure}
\hfill
\begin{subfigure}[b]{0.235\textwidth}
\centering
\includegraphics[scale=.33]{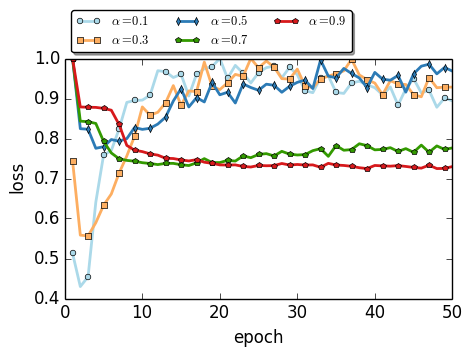} 
\caption{CSA loss}
\label{fig:CSA_loss_throughAlphas}
\end{subfigure}
\hfill
\caption{Classification vs. contrastive stance alignment (CSA) losses across training epochs and $\alpha$ values.}
\label{fig:Classification_CSA_losses_throughAlphas}
\vspace{-5pt}
\end{figure}

Figures~\ref{fig:Classification_loss_throughAlphas}~and~\ref{fig:CSA_loss_throughAlphas} show the classification ($L_{CA}$) versus contrastive stance ($L_{CSA}$) losses obtained from our best language adaptation model (i.e., without pretraining) across training epochs and $\alpha$ values. The results are averaged on validation data when performing $5$-fold cross-validation.
As Figure~\ref{fig:Classification_loss_throughAlphas} shows, there is greater reduction in the classification loss for smaller values of $\alpha$, i.e., when classification loss contributes more to the overall loss; see Equation~(\ref{eq:L_total}). On the other hand, Figure~\ref{fig:CSA_loss_throughAlphas} shows that 
the CSA loss decreases with larger values of $\alpha$ as the model pays more attention to the CSA loss; see the red and green lines in Figure~\ref{fig:CSA_loss_throughAlphas}. These results indicate that our language adaptation model can find a good balance between the classification loss and the CSA loss, with the value of $\alpha=.7$ yielding the best performance.

\begin{figure}[t]
\centering
\begin{subfigure}[b]{0.22\textwidth}
\centering
\includegraphics[scale=.33]{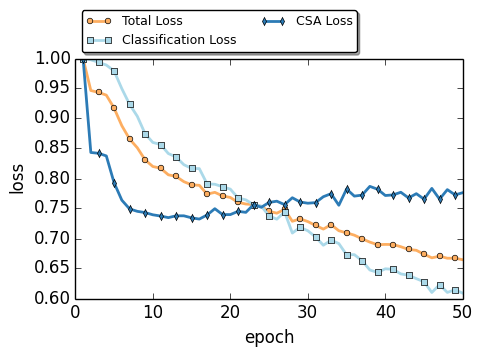}
\caption{without pretraining}
\label{fig:losses_without_pretraining}
\end{subfigure}
\hfill
\begin{subfigure}[b]{0.225\textwidth}
\centering
\includegraphics[scale=.33]{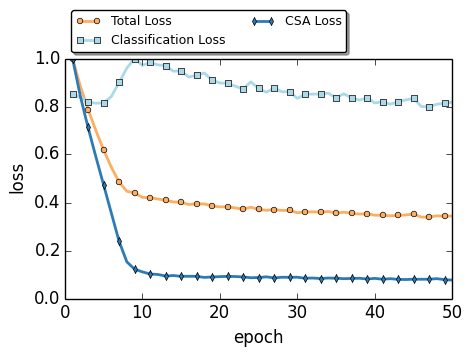}
\caption{with pretraining} 
\label{fig:losses_with_pretraining}
\end{subfigure}
\hfill
\caption{Classification loss vs. contrastive stance alignment loss (CSA) vs. total loss during training.}
\label{fig:compare_losses_w_wo_pretraining}
\vspace{-5pt}
\end{figure}

Figure 6 compares the classification $L_{CA}$, contrastive stance $L_{CSA}$, and total $L$ losses obtained by our CLMN model on the validation dataset across training epochs when the loss weight parameter ($\alpha$) is set to its best value. 
Figures~\ref{fig:losses_without_pretraining}~and~~\ref{fig:losses_with_pretraining} show the results without and with pretraining for $\alpha =.7$ and $\alpha=.3$ respectively.
Without pretraining (Figure~\ref{fig:losses_without_pretraining}), the classification (light-blue line) and CSA (dark-blue line) losses both decrease up to epoch $20$, after which the classification loss keeps decreasing, but the CSA loss starts increasing.
With pretraining (Figure~\ref{fig:losses_with_pretraining}), the CSA loss rapidly decreases for the first $10$ epochs (even though it has a small effect as $\alpha=.3$), and then continues with a smooth trend. This is because, during the initial training epochs, the model is biased to the {\em source} embedding space due to pretraining, and therefore the source and the target examples are far from each other. Then, our language adaptation model aligns the source and the target examples to form a much better shared embedding space and this alignment strategy yields a rapid decrease of the CSA loss in the first few epochs. Yet, in contrast to the CSA loss, the classification loss increases in the first few epochs.
This is because the model enforces alignment between the source and the target samples due to the large distances.
Finally, the total loss (orange line) indicates a good balance between the classification and the language adaptation losses, and it consistently decreases during training.

%% file: related_work.tex
\section{Related Work}
\label{sec:related_work}

\paragraph{Domain Adaptation.}
Previous work has presented several domain adaptation techniques. Unsupervised domain adaptation approaches~\cite{Ganin:2015:UDA:3045118.3045244,Long:2016:UDA:3157096.3157112,pmlr-v28-muandet13,conf/cvpr/GongSSG12} 
attempt to align the distribution of features in the embedding space mapped from the source and the target domains. 
A limitation of such approaches is that, even with perfect alignment, there is no guarantee that the same-label examples from different domains would map nearby in the embedding space. 
%
Supervised domain adaptation~\cite{Daume:2006:DAS:1622559.1622562,NIPS2013_5200,NIPS2010_4064} attempts to encourage same-label examples from different domains to map nearby in the embedding space. While supervised approaches perform better than unsupervised ones,
recent work~\cite{Motiian_2017_ICCV} has demonstrated superior performance by additionally encouraging class separation, meaning that examples from different domains and with different labels should be projected as far apart as possible in the embedding space.
Here, we combined both types of alignments
for stance detection.

\paragraph{Stance Detection.}
\citeauthor{mohammad2016semeval}~(\citeyear{mohammad2016semeval}) and \citeauthor{zarrella2016mitre}~(\citeyear{zarrella2016mitre}) worked on stances regarding target propositions, e.g., entities or events, as \emph{in-favor}, \emph{against}, or \emph{neither}. Most commonly, stance detection has been defined with respect to a \emph{claim} as \emph{agree}, \emph{disagree}, \emph{discuss} or \emph{unrelated}. 
Previous work mostly developed the models with rich hand-crafted features such as words, word embeddings, and sentiment lexicons~\cite{riedel2017simple,baird2017talos,Retrospective-C18-1158}. More recently, \citeauthor{mitra2018memory}~(\citeyear{mitra2018memory}) presented a mono-lingual and feature-light memory network for stance detection. 
In this paper, we built on this work to extend previous efforts in stance detection to a cross-lingual setting.

%% file: conclusion.tex
\section{Conclusion and Future Work}
\label{sec:conclusion}
We proposed an effective language adaptation approach to align class labels in source and target languages for accurate cross-lingual stance detection. Moreover, we investigated the behavior of our model in details and we have shown that it offers sizable performance gains over a number of competing approaches.
In future, we  will extend our language adaptation model to document retrieval and check-worthy claim detection tasks.
